\documentclass[letterpaper]{article} 
\usepackage{aaai23}  
\usepackage{times}  
\usepackage{helvet}  
\usepackage{courier}  
\usepackage[hyphens]{url}  
\usepackage{graphicx} 
\usepackage{natbib}  
\usepackage{caption} 
\usepackage{algorithm}
\usepackage{algorithmic}

\newcommand{\todo}[1]{{\color{red}\sf\bfseries [TODO]: #1}}

\newcommand{\mzou}[1]{{\color{blue}{[ZouMo: #1]}}}

\usepackage{amssymb} 
\usepackage{multirow}
\usepackage{color}
\usepackage{comment}
\usepackage{booktabs}
\usepackage{amsmath}

\usepackage{newfloat}
\usepackage{listings}
\DeclareCaptionStyle{ruled}{labelfont=normalfont,labelsep=colon,strut=off} 
\floatstyle{ruled}
\newfloat{listing}{tb}{lst}{}
\floatname{listing}{Listing}
\title{Simple and Efficient Heterogeneous Graph Neural Network}
\author{
    Xiaocheng Yang\textsuperscript{\rm 1},
    Mingyu Yan\footnote{Corresponding author.}\textsuperscript{\rm 1},
    Shirui Pan\textsuperscript{\rm 2},
    Xiaochun Ye\textsuperscript{\rm 1},
    Dongrui Fan\textsuperscript{\rm 1,3}
}
\affiliations{

    \textsuperscript{\rm 1} State Key Lab of Processors, Institute for Computing Technology, Chinese Academy of Sciences, China\\
    \textsuperscript{\rm 2} School of Information and Communication Technology, Griffith University, Australia\\
    \textsuperscript{\rm 3} School of Computer Science and Technology, University of Chinese Academy of Sciences, China\\
    \{yangxiaocheng,\,yanmingyu\}@ict.ac.cn,\,\,s.pan@griffith.edu.au,\,\,\{yexiaochun,\,fandr\}@ict.ac.cn
}

\usepackage{bibentry}

\begin{document}

\maketitle

\begin{abstract}
Heterogeneous graph neural networks (HGNNs) have powerful capability to embed rich structural and semantic information of a heterogeneous graph into node representations. 
Existing HGNNs inherit many mechanisms from graph neural networks (GNNs) designed for homogeneous graphs, especially the attention mechanism and the multi-layer structure.
These mechanisms bring excessive complexity, 
but seldom work studies whether they are really effective on heterogeneous graphs.
In this paper, we conduct an in-depth and detailed study of these mechanisms and
proposes \textit{Simple and Efficient Heterogeneous Graph Neural Network} (SeHGNN).
To easily capture structural information, SeHGNN pre-computes the neighbor aggregation using a light-weight mean aggregator, which reduces complexity by removing overused neighbor attention and avoiding repeated neighbor aggregation in every training epoch.
To better utilize semantic information, SeHGNN adopts the single-layer structure with long metapaths to extend the receptive field, as well as a transformer-based semantic fusion module to fuse features from different metapaths.
As a result, SeHGNN exhibits the characteristics of a simple network structure, high prediction accuracy, and fast training speed. Extensive experiments on five real-world heterogeneous graphs demonstrate SeHGNN's superiority over the state-of-the-arts on both accuracy and training speed.
\end{abstract}


\section{Introduction}



Recent years witness explosive growth in graph neural networks (GNNs) in pursuit of performance improvement of graph representation learning~\cite{comprehensive_gnn_survey,ijcai2022p772,lin2022comprehensive}. GNNs are primarily designed for homogeneous graphs associated with a single type of nodes and edges, following a neighborhood aggregation scheme to capture structural information of a graph, where the representation of each node is computed by recursively aggregating the features of neighbor nodes~\cite{kipf2016semi}.

\begin{figure}[ht]
  \centering
  \includegraphics[width=\linewidth]{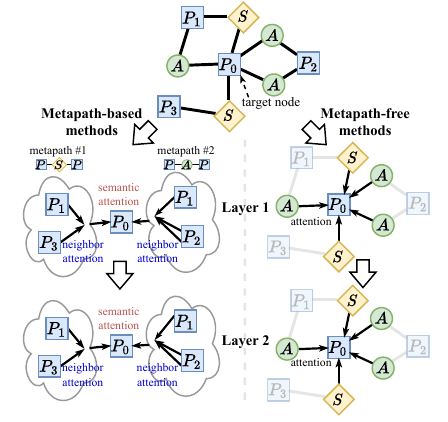}
  \caption{The general architectures of metapath-based methods and metapath-free methods on heterogeneous graphs.}
  \label{tab:hgnns_categories}
\end{figure}

However, GNNs are insufficient to deal with the heterogeneous graph which possesses rich semantic information in addition to structural information~\cite{HG_survey}. Many real-world data in complex systems are naturally represented as heterogeneous graphs, where multiple types of entities and relations among them are embodied by various types of nodes and edges, respectively. For example, as shown in Figure \ref{tab:hgnns_categories}, the citation network ACM includes several types of nodes: Paper (P), Author (A), and Subject (S), as well as many relations with different semantic meanings, such as Author$\xrightarrow{\rm writes}$Paper, Paper$\xrightarrow{\rm cites}$Paper, Paper$\xrightarrow{\rm belongs~to}$Subject.
These relations can be composited with each other to form high-level semantic relations, which are represented as metapaths~\cite{sun2011pathsim,sun2012mining}.
For example, the 2-hop metapath Author-Paper-Author (APA) represents the co-author relationship, 
while the 4-hop metapath Author-Paper-Subject-Paper-Author (APSPA) describes that the two authors have been engaged in the research of the same subject.
Heterogeneous graphs contain more comprehensive information and rich semantics and require specifically designed models.

Various heterogeneous graph neural networks (HGNNs) have been proposed to capture semantic information, achieving great performance in heterogeneous graph representation learning~\cite{hgnn_survey_tangjie,hgnn_survey_hanjiawei,hgnn_survey_shichuan,hgnn_survey_shiruipan,9855397}.
Therefore, HGNNs are at the heart of a broad range of applications such as social network analysis~\cite{liu2018heterogeneous}, recommendation~\cite{fan2019metapath,niu2020dual}, and knowledge graph inference~\cite{bansal2019a2n,vashishth2020compositionbased,wang2021mixed}.

Figure \ref{tab:hgnns_categories} depicts the two main categories of HGNNs.
\textit{Metapath-based methods}~\cite{RGCN2018, HetGNN2019, HAN2019, MAGNN2020} capture the structural information of the same semantic first and then fuse different semantic information. These models first aggregate neighbor features at the scope of each metapath to generate semantic vectors, and then fuse these semantic vectors to generate the final embedding vector.
\textit{Metapath-free methods}~\cite{RSHN2019, HetSANN, HGT, HGB} capture structural and semantic information simultaneously. These models aggregate messages from a node’s local neighborhood like traditional GNNs, but use extra modules (e.g. attentions) to embed semantic information such as node types and edge types into propagated messages.

Existing HGNNs inherit many mechanisms from GNNs over homogeneous graphs, especially the attention mechanism and the multi-layer structure, as illustrated in Figure \ref{tab:hgnns_categories}, but seldom work studies whether these mechanisms are really effective on heterogeneous graphs.
Moreover, the hierarchy attention calculation in the multi-layer network and repeated neighbor aggregation in every epoch bring excessive complexity and computation. For example, the neighbor aggregation process with attention modules takes more than 85\% of total time in the metapath-based model HAN~\cite{HAN2019} and the metapath-free model HGB~\cite{HGB}, which has become the speed bottleneck for the application of HGNNs on larger-scale heterogeneous graphs.

This paper provides an in-depth and detailed study of these mechanisms and yields two findings: (1) \textit{semantic attention is essential while neighbor attention is not necessary}, (2) \textit{models with a single-layer structure and long metapaths outperform those with multi-layers and short metapaths}. These findings imply that neighbor attention and multi-layer structure not only introduce unnecessary complexity, but also hinder models to from achieving better performance. 


To this end, we propose a novel metapath-based method named SeHGNN. SeHGNN employs the mean aggregator~\cite{sage233} to simplify neighbor aggregation, adopts a single-layer structure with long metapaths to extend the receptive field, and utilizes a transformer-based semantic fusion module to learn mutual attentions between semantic pairs. Additionally, as the simplified neighbor aggregation in SeHGNN is parameter-free and only involves linear operations, it earns the opportunity to execute the neighbor aggregation in the pre-processing step only once. As a result, SeHGNN not only demonstrates better performance but also avoids the need of repeated neighbor aggregation in every training epoch, leading to significant improvements in training speed.

We conduct experiments on four widely-used datasets from HGB benchmark~\cite{HGB} and a large-scale dataset from OGB challenge~\cite{OGB}. Results show that SeHGNN achieves superior performance over the state-of-the-arts for node classification on heterogeneous graphs.


The contributions of this work are summarized as follows:

\begin{itemize}
\item We conduct an in-depth study about the attention mechanism and the network structure in HGNNs and obtain two important findings, which reveal the needlessness of neighbor attention and the superiority of utilizing the single-layer structure and long metapaths.
\item Motivated by the findings above, we propose a simple and effective HGNN architecture SeHGNN. 
To easily capture structural information, SeHGNN pre-computes the neighbor aggregation in the pre-processing step using a light-weight mean aggregator, which removes the overused neighbor attention and avoids repeated neighbor aggregation in every training epoch.
To better utilize semantic information, SeHGNN adopts the single-layer structure with long metapaths to extend the receptive field, as well as a transformer-based semantic fusion module to fuse features from different metapaths.
\item Experiments on five widely-used datasets demonstrate the superiority of SeHGNN over the state-of-the-arts, i.e., high prediction accuracy and fast training speed.
\end{itemize}

\section{Preliminaries}






\textbf{Definition 1 \,\,\textit{Heterogeneous graphs}.\,}
\textit{A heterogeneous graph is defined as $G=\{V,E,\mathcal{T}^v,\mathcal{T}^e\}$, where $V$ is the set of nodes with a node type mapping function $\phi:V\rightarrow\mathcal{T}^v$, and $E$ is the set of edges with an edge type mapping function $\psi:E\rightarrow\mathcal{T}^e$.}
Each node $v_i{\in}V$ is attached with a node type $c_i{=}\phi(v_i){\in}\mathcal{T}^v$. Each edge $e_{t\leftarrow s}{\in}E\,$($e_{ts}$ for short) is attached with a relation $r_{c_t\leftarrow c_s}{=}\psi(e_{ts}){\in}\mathcal{T}^e$ ($r_{c_tc_s}$ for short), pointing from the source node $s$ to the target node $t$.
When $|\mathcal{T}^v|{=}|\mathcal{T}^e|{=}1$, the graph degenerates into homogeneous. 

The graph structure of $G$ can be represented by a series of adjacency matrices $\{A_r: r\in\mathcal{T}^e\}$. For each relation $r_{c_tc_s}{\in}\mathcal{T}^e$,\, $A_{c_tc_s}{\in}\mathbb{R}^{|V^{c_t}|\times |V^{c_s}|}$ is the corresponding adjacency matrix where the nonzero values indicate positions of edges $E^{c_tc_s}$ of the current relation.

\noindent \textbf{Definition 2 \,\,\textit{Metapaths}.\,}
\textit{A metapath defines a composite relation of several edge types, represented as $\mathcal{P}\triangleq c_1\leftarrow c_2\leftarrow \ldots\leftarrow c_{l}$\, ($\mathcal{P}= c_1c_2\ldots c_{l}$ for short).}

Given the metapath $\mathcal{P}$, a \textbf{metapath instance} $p$ is a node sequence following the schema defined by $\mathcal{P}$, represented as $p(v_1,v_l)=\{v_1,\,e_{12},\,v_2,\,\ldots,\,v_{l-1},\,e_{(l-1)l},\,v_l:v_i\in V^{c_i},\,e_{i(i+1)}\in E^{c_ic_{i+1}}\}$. In particular, $p(v_1,v_l)$ indicates the relationship of $(l{-}1)$-hop neighborhood where $v_1$ is the target node and $v_l$ is one of $v_1$'s \textbf{metapath-based neighbors}. 


Given a metapath $\mathcal{P}$ with the node types of its two ends as $c_1, c_l$, a \textbf{metapath neighbor graph} $G^\mathcal{P}=\{V^{c_1}\bigcup V^{c_l}, E^\mathcal{P}\}$ can be constructed out of $G$, where an edge $e_{ij}\in E^\mathcal{P},\phi(i)=c_1,\phi(j)=c_l$ exists in $G^\mathcal{P}$ if and only if there is an metapath instance $p(v_i,v_j)$ of $\mathcal{P}$ in $G$.

\section{Related Work} \label{sec:related_work}

For homogeneous graphs, GNNs are widely used to learn node representation from the graph structure. 
The pioneer GCN~\cite{kipf2016semi} proposes a multi-layer network following a layer-wise propagation rule, where the $l^{th}$ layer learns an embedding vector $h_v^{(l+1)}$ by aggregating features $\{h_u^{(l)}: u\in \mathcal{N}_v\}$ from the local 1-hop neighbor set $\mathcal{N}_v$ for each node $v\in V$.
GraphSAGE~\cite{sage233} improves the scalability for large graphs by introducing mini-batch training and neighbor sampling~\cite{9601152}.
GAT~\cite{velivckovic2017graph} introduces the attention mechanism to encourage the model to focus on the most important part of neighbors.
SGC~\cite{SGC} removes nonlinearities between consecutive graph convolutional layers,
which brings great acceleration and does not impact model effects.

Heterogeneous graphs contain rich semantics besides structural information, revealed by multiple types of nodes and edges. According to the way to deal with different semantics, HGNNs are categorized into metapath-based and metapath-free methods.

Metapath-based HGNNs first aggregate neighbor features of the same semantic and then fuse different semantics.
RGCN~\cite{RGCN2018} is the first to separate 1-hop neighbor aggregation according to edge types.
HetGNN~\cite{HetGNN2019} takes use of neighbors of different hops. It uses random walks to collect neighbors of different distances and then aggregates neighbors of the same node type.
HAN~\cite{HAN2019} utilizes metapaths to distinguish different semantics. It aggregates structural information with neighbor attention in each metapath neighbor graph in the \textit{neighbor aggregation} step, and then fuses outputs from different subgraphs with semantic attention for each node in the \textit{semantic fusion} step.
MAGNN~\cite{MAGNN2020} further leverages all nodes in a metapath instance rather than only the nodes of the two endpoints.

Metapath-free HGNNs aggregate messages from neighbors of all node types simultaneously within the local 1-hop neighborhood like GNNs, but using additional modules such as attentions to embed semantic information such as node types and edge types into propagated messages.
RSHN~\cite{RSHN2019} builds the coarsened line graph to obtain the global embedding representation of different edge types, and then it uses the combination of neighbor features and edge-type embeddings for feature aggregation in each layer.
HetSANN~\cite{HetSANN} uses a multi-layer GAT network with type-specific score functions to generate attentions for different relations.
HGT~\cite{HGT} proposes a novel heterogeneous mutual attention mechanism based on Transformer~\cite{transformer}, using type-specific trainable parameters for different types of nodes and edges.
HGB~\cite{HGB} takes the multi-layer GAT network as the backbone and incorporates both node features and learnable edge-type embeddings to generate attention values.

\begin{table}[!htbp]
\centering
\begin{tabular}{lcccc}
\hline
                      & \multicolumn{2}{c}{DBLP}      & \multicolumn{2}{c}{ACM}       \\ \hline
                      & macro-f1      & micro-f1      & macro-f1      & micro-f1      \\ \hline
HAN                   & 92.59         & 93.06         & 90.30         & 90.15         \\
HAN*                  & 92.75         & 93.23         & 90.61         & 90.48         \\
HAN${^\dagger}$       & 92.19         & 92.66         & 89.78         & 89.67         \\ \hline
HGB                   & 94.15         & 94.53         & 93.09         & 93.03         \\
HGB*                  & 94.20         & 94.58         & 93.11         & 93.05         \\
HGB${^\dagger}$       & 93.77         & 94.15         & 92.32         & 92.27         \\ \hline
\end{tabular}
\caption{Experiments to analyze the effects of two kinds of attentions. * means removing neighbor attention and $\dagger$ means removing semantic attention.} \label{tab:observation}
\end{table}

\begin{table}[ht]
\centering
\begin{tabular}{lcccc}
\hline
\multicolumn{1}{c}{}        & \multicolumn{2}{c}{DBLP} & \multicolumn{2}{c}{ACM} \\ \hline
\multicolumn{1}{c}{network} & macro-f1    & micro-f1   & macro-f1   & micro-f1   \\ \hline
(1,)                        & 79.43       & 80.16      & 89.81      & 90.03      \\ \hline
(1,1)                       & 85.06       & 86.69      & 90.79      & 90.87      \\
(2,)                        & 88.18       & 88.83      & 91.64      & 91.67      \\ \hline
(1,1,1)                     & 88.38       & 89.37      & 87.95      & 88.84      \\
(3,)                        & 93.33       & 93.72      & 92.67      & 92.64      \\ \hline
(1,1,1,1)                   & 89.55       & 90.44      & 88.62      & 88.93      \\
(2,2)                       & 91.88       & 92.35      & 92.57      & 92.53      \\
(4)                         & \textbf{93.60}       & \textbf{94.02}      & \textbf{92.82}      & \textbf{92.79}      \\ \hline
\end{tabular}
\caption{Experiments to analyze the effects of different combinations of the number of layers and the maximum metapath hop. e.g., the structure (1,1,1) means a three-layer network with all metapaths no more than 1 hop in each layer.}
\label{tab:observation2}
\end{table}

Except for the above two main categories of HGNNs, SGC-based work such as NARS~\cite{NARS2020}, SAGN~\cite{SAGN2021}, and GAMLP~\cite{GAMLP2021} also show impressive results on heterogeneous graphs, but they aggregate features of all types of nodes together without explicitly distinguishing different semantics.

\section{Motivation} \label{sec:motivation}


\begin{figure*}[ht]
  \centering
  \includegraphics[width=\linewidth]{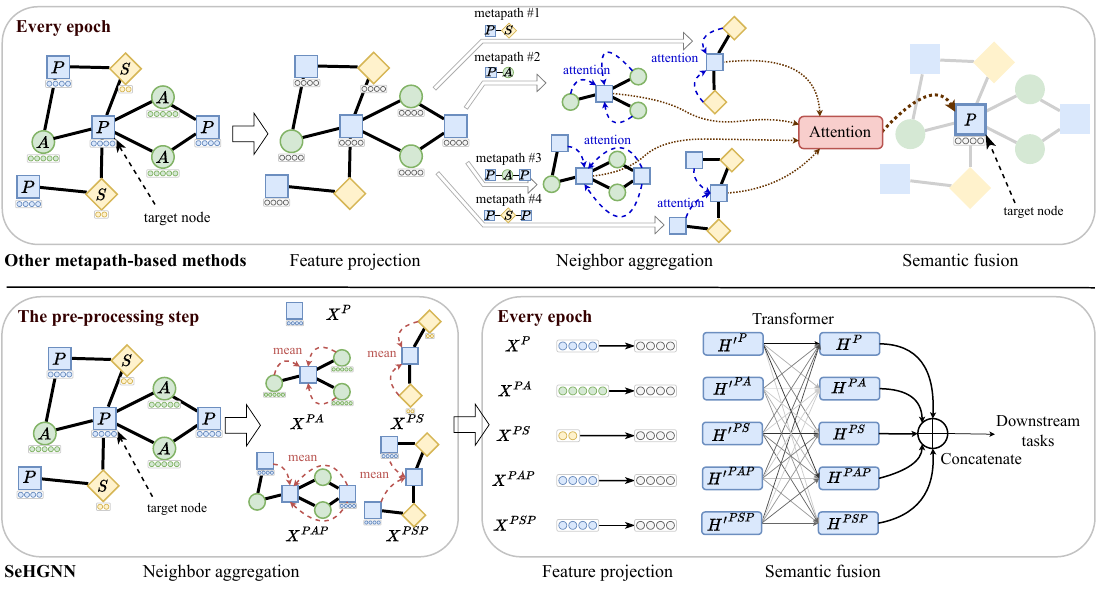}
  \caption{The architecture of SeHGNN compared to previous metapath-based methods. The example is based on ACM dataset with node types author (A), paper (P), and subject (S). This figure exhibits aggregation of 0-hop metapath P (the target node itself), 1-hop metapaths PA, PS, and 2-hop metapaths PAP, PSP.}
  \label{tab:framework}
\end{figure*}

Existing HGNNs inherit many mechanisms from GNNs without analysis of their effects.
In this section, we conduct an in-depth and detailed study of widely-used mechanisms, i.e. the attention mechanism and the multi-layer structure for HGNNs. Through experiments, we obtain two important findings that guide us in designing the architecture of SeHGNN. All results presented in this section are the average of 20 runs with different data partitions to mitigate the influence of random noise.

\noindent \textbf{Study on attentions.}
HGNNs use multiple attentions, as shown in Figure~\ref{tab:hgnns_categories}, which are calculated using distinct modules or parameters. These attentions can be classified into two types: \textit{neighbor attention} within neighbors of the same relation and \textit{semantic attention} among different relations. Metapath-based methods like HAN incorporate two attentions in the neighbor aggregation step and the semantic fusion step, respectively. Metapath-free methods like HGB compute attentions of 1-hop neighbors with relation-specific embeddings. While distinguishing between the two types of attention in metapath-free methods can be challenging, we can perform additional calculations to eliminate the influence of either attention. Specifically, for the attentions values of each node's neighbors, we can average them within each relation which equals removing neighbor attention, or normalize them within each relation so that each relation contributes equally to final results to remove semantic attention.


Re-implement experiments on HGB reveal that a well-trained HGB model tends to assign similar attention values within each relation, leading us to investigate the necessity of different attentions. We conduct experiments on HAN and HGB, where * means removing neighbor attention and ${^\dagger}$ means removing semantic attention. Results in Table \ref{tab:observation} indicate that models without semantic attention exhibit a decrease of model effects, while models without neighbor attention do not, from which we obtain the first finding.

\textit{Finding 1: Semantic attention is essential, while neighbor attention is not necessary.} This finding is reasonable, as semantic attention is able to weigh the importance of different semantics. And for neighbor attention, various SGC-based work~\cite{SGC,rossi2020sign,GAMLP2021} has demonstrated that simple mean aggregation can be just as effective as aggregation with attention modules.

\noindent \textbf{Study on multi-layer structure.}
Without neighbor attention, metapath-free methods have an equivalent form that first averages features of neighbors within each relation and then fuses outputs of different relations. Therefore, they can be converted to metapath-based methods with a multi-layer structure and only 1-hop metapaths in each layer. So, in the following experiments, we focus on the influence of number of layers and metapaths in metapath-based methods.

We conduct experiments on HAN using a list of numbers to represent the structure of each variant. For instance, on the ACM dataset, the structure (1,1,1,1) represents a four-layer network with 1-hop metapaths PA, PS in each layer, and (4) represent a single-layer network with all metapaths no more than 4-hop, such as PA, PS, PAP, PSP, PAPAP, PSPSP and so on. These lists also exhibit the sizes of receptive field. For example, structures (1,1,1,1), (2,2), (4) have the same receptive field size, which involves 4-hop neighbors. 
Based on the results shown in Table \ref{tab:observation2}, we conclude the second finding.

\textit{Finding 2: Models with a single-layer structure and long metapaths outperform those with multi-layers and short metapaths.} As shown in Table \ref{tab:observation2}, models with a single-layer and long metapaths achieve better performance under the same size of the receptive field. 
We attribute this to the fact that multi-layer networks fuse semantics per layer, making it difficult to distinguish high-level semantics. For instance, in a model with the structure (4), the utilization of multi-hop metapaths allow us to distinguish between high-level semantics, such as being written by the same author (PAP) or familiar authors (PAPAP), whereas these distinctions are unavailable in a four-layer network (1,1,1,1) as all intermediate vectors between two consecutive layers represent mixtures of different semantics.
Moreover, increasing the maximum metapath length enhances the model's performance by introducing more metapaths with different semantics.

\noindent \textbf{Proposal of SeHGNN.}
Motivated by the two findings, on one hand, we can avoid redundant neighbor attention by employing mean aggregation at the scope of each metapath without sacrificing model effects; on the other hand, we can simplify the network structure using a single layer, but use more and longer metapaths to expand the receptive field and achieve better performance. Furthermore, as the neighbor aggregation part without the attention module involves only linear operations and no trainable parameters, it endows an opportunity to execute neighbor aggregation in the pre-processing step only once rather than in every training epoch, which significantly reduces the training time. Overall, these optimizations simplify the network structure and make it more efficient, which are the key points of SeHGNN.

\section{Methodology}

This section formally proposes \textit{Simple and Efficient Heterogeneous Neural Network} (SeHGNN).
SeHGNN's architecture is illustrated in Figure \ref{tab:framework} and includes three primary components: simplified neighbor aggregation, multi-layer feature projection, and transformer-based semantic fusion. 
Figure \ref{tab:framework} also highlights the distinction between SeHGNN and other metapath-based HGNNs, i.e., SeHGNN pre-computes the neighbor aggregation in the pre-processing step, thereby avoiding the excessive complexity of repeated neighbor aggregation in every training epoch.
Algorithm \ref{alg:algorithm} outlines the overall training process.

\subsection{Simplified Neighbor Aggregation} \label{sec:simplified_neighbor_aggregation}


The simplified neighbor aggregation is executed only once in the pre-processing step and generates a list of feature matrices $M=\{X^\mathcal{P}:\mathcal{P}\in\Phi_X\}$ of different semantics for the set $\Phi_X$ of all given metapaths. 
Generally, for each node $v_i$, it employs the mean aggregation to aggregate features from metapath-based neighbors for each given metapath, and outputs a list of semantic feature vectors, denoted as
$$
m_i=\{\mathbf{z}^\mathcal{P}_i=\frac{1}{||S^\mathcal{P}||}\sum_{p(i,j)\in S_\mathcal{P}}\mathbf{x}_j:\mathcal{P}\in\Phi_X\}, \label{intra-aggr-node}
$$
where $S^\mathcal{P}$ is the set of all metapath instances corresponding to metapath $\mathcal{P}$ and $p(i,j)$ represents one metapath instance with the target node $i$ and the source node $j$.

We propose a new method to simplify the collection of metapath-based neighbors.
Existing metapath-based methods like HAN build metapath neighbor graphs that enumerate all metapath-based neighbors for each metapath, which brings a high overhead as the number of metapath instances grows exponentially with the length of the metapath. Inspired by the layer-wise propagation of GCN, we calculate the final contribution weight of each node to targets using the multiplication of adjacency matrices. Specifically, let $X^c=\{{x_0^c}^T;{x_1^c}^T;$ $\ldots;{x_{||V^c||-1}^c}^T\}\in\mathbb{R}^{||V^c||\times d^c}$ be the raw feature matrix of all nodes belonging to type $c$, where
$d^c$ is the feature dimension, and then the simplified neighbor aggregation process can be expressed as
$$
X^\mathcal{P}=\hat{A}_{c,c_1}\hat{A}_{c_1,c_2}\ldots\hat{A}_{c_{l-1},c_l}X^{c_l}, \label{intra-aggr-matrix}
$$
where $\mathcal{P}=cc_1c_2\ldots c_l$ is a $l$-hop metapath, and $\hat{A}_{c_i,c_{i+1}}$ is the row-normalized form of adjacency matrix $A_{c_i,c_{i+1}}$ between node type $c_i$ and $c_{i+1}$.
Please note that, the aggregation results of short metapaths can be used as intermediate values for long metapaths. 
For example, given two metapaths PAP and PPAP for the ACM dataset, we can calculate $X^{PAP}$ first and then calculate $X^{PPAP}=\hat{A}_{PP}X^{PAP}$.

\begin{algorithm}[tb]
\caption{The overall training process of SeHGNN}
\label{alg:algorithm}
\textbf{Input}: Raw feature matrices $\{X^{c_i}: c_i\in \mathcal{T}^v\}$; the raw label matrix $Y$; metapath sets $\Phi_X$ for features and $\Phi_Y$ for labels\\
\textbf{Parameter}: ${\rm MLP}_{\mathcal{P}_i}$ for feature projection; $W_Q$, $W_K$, $W_V$, $\beta$ for semantic fusion; ${\rm MLP}$ for downstream tasks \\
\textbf{Output}: Node classification results ${\rm Pred}$ for target type $c$

\begin{algorithmic}[1] 
\STATE \textbf{\% Neighbor aggregation}
\STATE Calculate aggregation of raw features for each $\mathcal{P}\in\Phi_X$ \\ $X^\mathcal{P}{=}\hat{A}_{c,c_1}\ldots \hat{A}_{c_{l{-}1}c_l}X^{c_l},\,\,\,\,\mathcal{P}{=}cc_1\ldots c_{l{-}1}c_l$
\STATE Calculate aggregation of labels for each $\mathcal{P}\in\Phi_Y$ \\
$Y^\mathcal{P}{=}{\rm rm\_diag}(\hat{A}_{c,c_1}\ldots \hat{A}_{c_{l{-}1}c})Y,\,\,\,\,\mathcal{P}{=}cc_1\ldots c_{l{-}1}c$
\STATE Collect all semantic matrices \\ $M=\{X^\mathcal{P}:\mathcal{P}\in\Phi_X\}\bigcup\{Y^\mathcal{P}:\mathcal{P}\in\Phi_Y\}$
\FOR{${\rm each\,epoch}$}
\STATE \textbf{\% Feature projection}
\STATE ${H'}^{\mathcal{P}_i}={\rm MLP}_{\mathcal{P}_i} (M^{\mathcal{P}_i}),\,\,M^{\mathcal{P}_i}\in M$
\STATE \textbf{\% Semantic fusion}
\STATE $Q^{\mathcal{P}_i}{=}W_Q{H'}^{\mathcal{P}_i},\,K^{\mathcal{P}_i}{=}W_K{H'}^{\mathcal{P}_i},\,V^{\mathcal{P}_i}{=}W_V{H'}^{\mathcal{P}_i}$
\STATE $\alpha_{ij}=\frac{\exp{(Q^{\mathcal{P}_i}\cdot {K^{\mathcal{P}_j}}^T})}{\sum_{t}\exp{(Q^{\mathcal{P}_i}\cdot {K^{\mathcal{P}_t}}^T)}}$
\STATE $H^{\mathcal{P}_i}=\beta \sum_j\alpha_{ij}V^{\mathcal{P}_i} + {H'}^{\mathcal{P}_i}$
\STATE $H^c={\rm concatenate}([H^{\mathcal{P}_1} || H^{\mathcal{P}_2} || \ldots])$
\STATE \textbf{\% Downstream tasks}
\STATE ${\rm Pred}={\rm MLP}(H^c)$
\STATE Calculate loss function $\mathcal{L}=-\sum_i y_i\ln pred_i$, do back-propagation and update network parameters
\ENDFOR
\STATE \textbf{return} ${\rm Pred}$
\end{algorithmic}
\end{algorithm}

Besides, previous~\cite{wang2020unifying, wang2021bag, shi2020masked} research has demonstrated that incorporating labels as additional inputs can improve model performance. To capitalize on this, similar to the aggregation of raw features, we represent labels in the one-hot format and propagate them across various metapaths.
This process generates a series of matrices $\{Y^\mathcal{P}:\mathcal{P}\in\Phi_Y\}$ that reflect the label distribution of corresponding metapath neighbor graphs.
Please note that the two endpoints of any metapath $\mathcal{P}\in\Phi_Y$ should be the target node type $c$ in the node classification task.
Given a metapath $\mathcal{P}=cc_1c_2\ldots c_{l-1}c\in\Phi_Y$,
the label propagation process can be represented as
$$
Y^\mathcal{P}={\rm rm\_diag}(\hat{A}^\mathcal{P})Y^c,\, \hat{A}^\mathcal{P}=\hat{A}_{c,c_1}\hat{A}_{c_1,c_2}\ldots\hat{A}_{c_{l-1},c},
$$
where $Y^c$ is the raw label matrix. In the matrix $Y^c$, rows corresponding to nodes in the training set take the values of one-hot format labels, while other rows are filled with 0. To avoid label leakage, we prevent each node from receiving the ground truth label information of itself, by removing the diagonal values in the results of multiplication of adjacency matrices. The label propagation also executes in the neighbor aggregation step and produces semantic matrices as extra inputs for later training.

\begin{table*}[!htbp]
\centering\small
\begin{tabular}{cccccccccc}
\hline
                     &           & \multicolumn{2}{c}{DBLP}                  & \multicolumn{2}{c}{IMDB}                  & \multicolumn{2}{c}{ACM}                   & \multicolumn{2}{c}{Freebase}              \\ \hline
                     &           & macro-f1            & micro-f1            & macro-f1            & micro-f1            & macro-f1            & micro-f1            & macro-f1            & micro-f1            \\ \hline
\multirow{4}{*}{1st} & RGCN      & 91.52±0.50          & 92.07±0.50          & 58.85±0.26          & 62.05±0.15          & 91.55±0.74          & 91.41±0.75          & 46.78±0.77          & 58.33±1.57          \\
                     & HetGNN    & 91.76±0.43          & 92.33±0.41          & 48.25±0.67          & 51.16±0.65          & 85.91±0.25          & 86.05±0.25          & -                   & -                   \\
                     & HAN       & 91.67±0.49          & 92.05±0.62          & 57.74±0.96          & 64.63±0.58          & 90.89±0.43          & 90.79±0.43          & 21.31±1.68          & 54.77±1.40          \\
                     & MAGNN     & 93.28±0.51          & 93.76±0.45          & 56.49±3.20          & 64.67±1.67          & 90.88±0.64          & 90.77±0.65          & -                   & -                   \\ \hline
\multirow{4}{*}{2nd} & RSHN      & 93.34±0.58          & 93.81±0.55          & 59.85±3.21          & 64.22±1.03          & 90.50±1.51          & 90.32±1.54          & -                   & -                   \\
                     & HetSANN   & 78.55±2.42          & 80.56±1.50          & 49.47±1.21          & 57.68±0.44          & 90.02±0.35          & 89.91±0.37          & -                   & -                   \\
                     & HGT       & 93.01±0.23          & 93.49±0.25          & 63.00±1.19          & 67.20±0.57          & 91.12±0.76          & 91.00±0.76          & 29.28±2.52          & 60.51±1.16          \\
                     & HGB       & 94.01±0.24          & 94.46±0.22          & 63.53±1.36          & 67.36±0.57          & 93.42±0.44          & 93.35±0.45          & 47.72±1.48          & \textbf{66.29±0.45} \\ \hline
3rd                  & SeHGNN    & \textbf{95.06±0.17} & \textbf{95.42±0.17} & \textbf{67.11±0.25} & \textbf{69.17±0.43} & \textbf{94.05±0.35} & \textbf{93.98±0.36} & \textbf{51.87±0.86} & 65.08±0.66          \\ \hline
\multirow{4}{*}{4th} & Variant\#1 & 93.61±0.51 & 94.08±0.48 & 64.48±0.45                   & 66.58±0.42                   & 93.06±0.18                   & 92.98±0.18                   & 33.23±1.39                   & 57.60±1.17                   \\
                     & Variant\#2 & 94.66±0.27                   & 95.01±0.24                   & 65.27±0.60                   & 66.68±0.52      & 93.46±0.43                   & 93.38±0.44                                & 46.82±1.12                   & 64.08±1.43                   \\
                     & Variant\#3 & 94.86±0.14          & 95.24±0.13          & 66.63±0.34          & 68.21±0.32          & 93.95±0.48          & 93.87±0.50          & 50.71±0.44          & 63.41±0.47          \\
                     & Variant\#4 & 94.52±0.05          & 94.93±0.06          & 64.99±0.54          & 66.65±0.50          & 93.88±0.63          & 93.80±0.64          & 50.30±0.23          & 64.53±0.38          \\ \hline
\end{tabular} 
\caption{Experiment results on the four datasets from HGB benchmark, where ``-'' means that the models run out of memory.}
\label{tab:result_on_midlle_dataset}
\end{table*}

\subsection{Multi-layer Feature Projection} \label{sec:multi_layer_feature_projection}

The feature projection step projects semantic vectors into the same data space, as semantic vectors of different metapaths may have different dimensions or lie in various data spaces. Generally, it defines a semantic-specific transformation matrix $W^\mathcal{P}$ for each metapath $\mathcal{P}$ and calculates ${H'}^\mathcal{P}=W^\mathcal{P}X^\mathcal{P}$.
For better representation power, we use a multi-layer perception block $\rm{MLP}_\mathcal{P}$ for each metapath $\mathcal{P}$ with a normalization layer, a nonlinear layer, and a dropout layer between two consecutive linear layers. Then, this process can be expressed as
$${H'}^\mathcal{P}=\rm{MLP}_\mathcal{P}(X^\mathcal{P}).$$

\subsection{Transformer-based Semantic Fusion} \label{sec:semantic_aggregation}

The semantic fusion step fuses semantic feature vectors and generates the final embedding vector for each node. Rather than using a simple weighted sum format, we propose a transformer~\cite{transformer}\,-based semantic fusion module to further explore the mutual relationship between each pair of semantics.



The transformer-based semantic fusion module is designed to learn the mutual attention between pairs of semantic vectors, given the pre-defined metapath list $\Phi=\{\mathcal{P}_1,\ldots,\mathcal{P}_K\}$ and projected semantic vectors $\{{h'}^{\mathcal{P}_1},\ldots,{h'}^{\mathcal{P}_K}\}$ for each node.
For each semantic vector ${h'}^{\mathcal{P}_i}$, the module maps this vector into a query vector $q^{\mathcal{P}_i}$, a key vector $k^{\mathcal{P}_i}$, and a value vector $v^{\mathcal{P}_i}$. The mutual attention weight $\alpha_{(\mathcal{P}_i,\mathcal{P}_j)}$ is the dot product result of the query vector $q^{\mathcal{P}_i}$ and the key vector $k^{\mathcal{P}_j}$ after a softmax normalization. The output vector $h^{\mathcal{P}_i}$ of current semantic $\mathcal{P}_i$ is the weighted sum of all value vectors $v^{\mathcal{P}_j}$ plus a residual connection. The process of semantic fusion can be presented as
$$
q^{\mathcal{P}_i}=W_Q{h'}^{\mathcal{P}_i},\,
k^{\mathcal{P}_i}=W_K{h'}^{\mathcal{P}_i},\,
v^{\mathcal{P}_i}=W_V{h'}^{\mathcal{P}_i},\,\,\mathcal{P}_i\in\Phi,
$$
$$
\alpha_{(\mathcal{P}_i,\mathcal{P}_j)}=\frac{\exp{(q^{\mathcal{P}_i}\cdot {k^{\mathcal{P}_j}}^T)}}{\sum_{\mathcal{P}_t\in\Phi}\exp{(q^{\mathcal{P}_i}\cdot {k^{\mathcal{P}_t}}^T)}},
$$
$$
h^{\mathcal{P}_i}=\beta\sum_{\mathcal{P}_j\in\Phi}\alpha_{(\mathcal{P}_i,\mathcal{P}_j)}\,v^{\mathcal{P}_j}+{h'}^{\mathcal{P}_i},
$$
where $W_Q,W_K,W_V,\beta$ are trainable parameters shared across all metapaths.

The final embedding vector of each node is the concatenation of all those output vectors. For downstream tasks like the node classification, another MLP is used to generate prediction results, which can be expressed as
$$
{\rm Pred}={\rm MLP}([h^{\mathcal{P}_1} || h^{\mathcal{P}_2} || \ldots || h^{\mathcal{P}_{|\Phi|}}]).
$$

\section{Experiment}







Experiments are conducted on four widely-used heterogeneous graphs including DBLP, ACM, IMDB, and Freebase from HGB benchmark~\cite{HGB}, as well as a large-scale dataset ogbn-mag from OGB challenge~\cite{hu2021ogb}. The details about all experiment settings and the network configurations are recorded in Appendix\footnote{Appendix can be found at \url{https://arxiv.org/abs/2207.02547}. Codes are available at \url{https://github.com/ICT-GIMLab/SeHGNN}.}.

\subsection{Results on HGB Benchmark}

Table~\ref{tab:result_on_midlle_dataset} presents the performance of SeHGNN on four datasets compared to several baselines in HGB benchmark, including four metapath-based methods (1st block) and four metapath-free methods (2nd block). 
Results demonstrate the effectiveness of SeHGNN as it achieves the best performance over all these baselines but the second best for micro-f1 accuracy on the Freebase dataset.

Additionally, we conduct comprehensive ablation studies to validate the two findings in the Motivation section and to determine the importance of other modules. The 4th block of Table~\ref{tab:result_on_midlle_dataset} shows the results of four variants of SeHGNN.

Variant\#1 utilizes GAT for each metapath in the neighbor aggregation step like HAN. Variant\#2 uses a two-layer structure, where each layer has independent neighbor aggregation and semantic fusion steps, but the maximum hop of metapaths in each layer is half of that in SeHGNN to ensure that SeHGNN and its Variant\#2 have the same size of receptive field. The performance gap between SeHGNN and its two variants proves that the two findings also apply for SeHGNN. 

Variant\#3 does not include labels as extra inputs, and Variant\#4 replaces the transformer-based semantic fusion with the weighted sum fusion like HAN. Notably, although inferior to SeHGNN, Variant\#3 has already outperformed most baselines except the micro-f1 on the Freebase dataset. These results show that the utilization of label propagation and transformer-based fusion improves model performance.

\begin{table}[!t]
\centering\small
\begin{tabular}{lcc}
\hline
Methods               & Validation accuracy & Test accuracy       \\ \hline
RGCN                  & 48.35±0.36          & 47.37±0.48          \\
HGT                   & 49.89±0.47          & 49.27±0.61          \\
NARS                  & 51.85±0.08          & 50.88±0.12          \\
SAGN                  & 52.25±0.30          & 51.17±0.32          \\
GAMLP                 & 53.23±0.23          & 51.63±0.22          \\ \hline
HGT+emb               & 51.24±0.46          & 49.82±0.13          \\
NARS+emb              & 53.72±0.09          & 52.40±0.16          \\
GAMLP+emb             & 55.48±0.08          & 53.96±0.18          \\
SAGN+emb+ms           & 55.91±0.17          & 54.40±0.15          \\
GAMLP+emb+ms          & 57.02±0.41          & 55.90±0.27          \\ \hline
SeHGNN                & 55.95±0.11          & 53.99±0.18          \\
SeHGNN+emb            & 56.56±0.07          & 54.78±0.17          \\
SeHGNN+ms             & 58.70±0.08          & 56.71±0.14          \\
SeHGNN+emb+ms         & \textbf{59.17±0.09} & \textbf{57.19±0.12} \\ \hline
\end{tabular} 
\caption{Experiment results on ogbn-mag compared with methods on the OGB leaderboard, where ``emb'' means using extra embeddings and ``ms'' means multi-stage training.}
\label{tab:result_on_large_dataset}
\end{table}

\subsection{Results on Ogbn-mag}

 The ogbn-mag dataset presents two extra challenges: (1) some types of nodes lack raw features, (2) target type nodes are split according to years, causing training nodes and test nodes to have different data distribution. Existing methods usually address these challenges by (1) generating extra embeddings (abbreviated as \textit{emb}) using unsupervised representation learning algorithms like ComplEx~\cite{ComplEx2016} and (2) utilizing multi-stage learning (abbreviated as \textit{ms}), which selects test nodes with confident predictions in last training stage, adds these nodes to the training set and re-trains the model in the new stage~\cite{li2018deeper, sun2020multi, yang2021self}. To provide a comprehensive comparison, we compare results with or without these tricks. For methods without \textit{emb}, we use randomly initialized raw feature vectors.

Table~\ref{tab:result_on_large_dataset} displays the results on the large-scale dataset ogbn-mag compared with baselines on the OGB leaderboard. Results show that SeHGNN outperforms other methods under the same condition. It is worth noting that SeHGNN with randomly initialized features even outperforms others with well-trained embeddings from additional representation learning algorithms, which reflects that SeHGNN learns more information from the graph structure.

\subsection{Time Analysis}
Firstly, we theoretically analyze the time complexity of SeHGNN compared to HAN and HGB as Table~\ref{tab:time_complexity} shows. We assume a one-layer structure with $k$ metapaths for SeHGNN and HAN, and $l$-layer structure for HGB. The maximum hop of metapaths is also $l$ to ensure the same size of receptive field. The number of target type nodes\footnote{For concise comparison, we only consider the complexity of linear projection for target type nodes on methods HAN and HGB. Please refer to Appendix for details.} is $n$ and the dimension of input and hidden vectors is $d$. The average number of neighbors in metapath neighbor graphs on HAN and involved neighbors during multi-layer aggregation on HGB are $e_1, e_2$, respectively. Please note that both $e_1$ and $e_2$ grow exponentially with the length of metapaths and layer number $l$. For the above five datasets we use tens of metapaths at most, but each node averagely aggregates information from thousands of neighbors for $l\ge 3$. Generally, we have $e_1\gg k^2, e_2 \gg k^2$, so the theoretical complexity of SeHGNN is much lower than that of HAN and HGB.

\begin{figure}[!t]
  \centering
  \includegraphics[width=\linewidth]{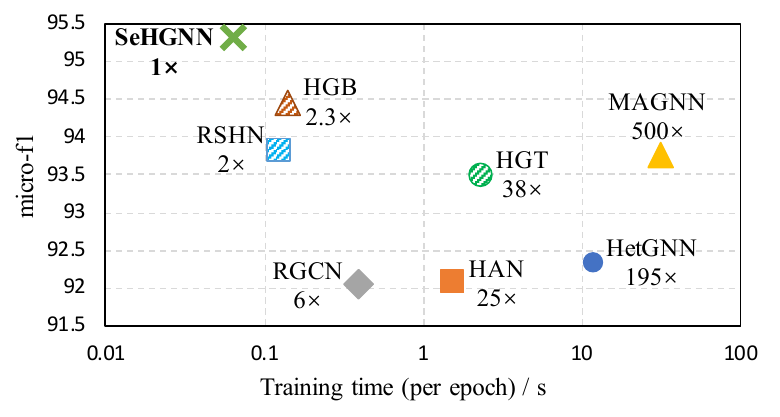}
  \caption{Micro-f1 scores and time consumption of different HGNNs on DBLP dataset. Numbers below model names exhibit the ratio of time consumption relative to SeHGNN. e.g., ``6x'' below RGCN means RGCN costs 6 times of time.} \label{fig:time_usage}
\end{figure}

To validate our theoretical analysis, we conduct experiments to compare the time consumption of SeHGNN with previous HGNNs. Figure~\ref{fig:time_usage} shows achieving micro-f1 scores relative to the average time consumption of each training epoch for these models, which reflects the superiority of SeHGNN on both the training speed and the model effect.


\begin{table}[!t]
\small
\tabcolsep=-1pt
\begin{tabular}{ccccc}
\hline
       & \begin{tabular}[c]{@{}c@{}}Feature\\ projection\end{tabular} & \begin{tabular}[c]{@{}c@{}}Neighbor\\ \,\,\,aggregation\end{tabular} & \begin{tabular}[c]{@{}c@{}}Semantic\\ fusion\end{tabular} & Total \\ \hline
\,SeHGNN & $\,\,O(nkd^2)$ & -- & $O(n(kd^2{+}k^2d))$ & $\,\,O(nd(k^2{+}kd))$ \\
HAN    & $O(nd^2)$ & $O(nke_1d)$ & $O(nkd^2)$ & $O(nd(ke_1{+}kd))$ \\
HGB    & $O(nld^2)$ & \multicolumn{2}{c}{$O(ne_2d)$\,\,\,\,\,\,\,\,} & $O(nd(e_2{+}ld))$ \\ \hline
\end{tabular} 
\caption{Time complexity of SeHGNN, HAN and HGB.}
\label{tab:time_complexity}
\end{table}

\section{Conclusion}


This paper proposes a novel approach called SeHGNN for heterogeneous graph representation learning, which is based on two key findings about attention utilization and network structure.
SeHGNN adopts a light-weight mean aggregator to pre-compute neighbor aggregation, which effectively captures structural information while avoiding overused neighbor attention and repeated neighbor aggregation.
Moreover, SeHGNN utilizes a single-layer structure with long metapaths to extend the receptive field and a transformer-based semantic fusion module to better utilize semantic information, resulting in significant improvements in model effectiveness. 
Experiments on five commonly used datasets demonstrate that SeHGNN outperforms state-of-the-art methods in terms of both accuracy and training speed.

\section{Acknowledgments}
This work was supported by the National Natural Science Foundation of China (Grant No. 61732018, 61872335, and 62202451), Austrian-Chinese Cooperative R\&D Project (FFG and CAS) (Grant No. 171111KYSB20200002), CAS Project for Young Scientists in Basic Research (Grant No. YSBR-029), and CAS Project for Youth Innovation Promotion Association.

\bibliography{aaai23}

\appendix
\section{Appendix}

\subsection{Observation in HGB models}

The metapath-free method HGB~\cite{HGB} calculates the attention value over each edge using the concatenation of node embeddings of the two endpoints and the edge-type embedding, presented as
{\small
$$\alpha_{ij}=\frac{{\rm exp}({\rm LeakyReLU}(\mathbf{a}^T[\mathbf{Wh}_i||\mathbf{Wh}_j||\mathbf{W}_r\mathbf{r}_{ij}]))}{\sum_{k\in\mathcal{N}_i}{\rm exp}({\rm LeakyReLU}(\mathbf{a}^T[\mathbf{Wh}_i||\mathbf{Wh}_k||\mathbf{W}_r\mathbf{r}_{ik}]))}.$$
}

\begin{figure}[!b]
  \centering
  \includegraphics[width=\linewidth]{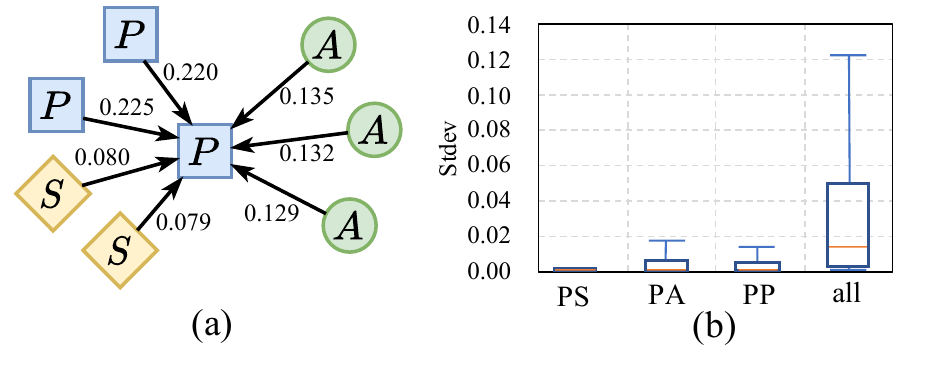}
  \caption{(a) An illustration of the observation that HGB tends to assign similar attention values within each relation for each target node. (b) The box plot of standard deviations of attention values for different relations in HGB.}
  \label{tab:stdev}
\end{figure}

\begin{table*}[!ht]
\tabcolsep=3.5pt
\small\centering
\begin{tabular}{|c|c|c|cc|}
\hline
                                                                                & RGCN & HetGNN & \multicolumn{1}{c|}{HAN} & MAGNN \\ \hline
\begin{tabular}[c]{@{}c@{}}Feature\\ projection\end{tabular}                     & $\textbf{h}_v^r=W^{\psi(r)} \textbf{x}_v$ & $\textbf{h}'_v={\rm Bi\raisebox{0mm}{-}LSTM}^{\phi(v)} (\textbf{x}_v)$ & \multicolumn{2}{c|}{$\textbf{h}'_v=W^{\phi(v)} \textbf{x}_v$}                                                                                                                                                                        \\ \hline
\multirow{3}{*}{\begin{tabular}[c]{@{}c@{}}Neighbor\\ aggregation\end{tabular}} & {\multirow{3}{*}{$\textbf{z}_v^r=\frac{1}{|\mathcal{N}_v^r|}\sum_{u\in\mathcal{N}_v^r}\textbf{h}^r_u$}} & {\multirow{3}{*}{$\textbf{z}_v^t={\rm Bi\raisebox{0mm}{-}LSTM}(\{\textbf{h}'_u: u\in \mathcal{N}_v^t\})$}}                             & \multicolumn{1}{c|}{$\gamma^\mathcal{P}_{v,u}=\sigma(\textbf{a}^T_{\mathcal{P}}\cdot [\textbf{h}'_v || \textbf{h}'_u])$}    & \begin{tabular}[c]{@{}c@{}}$\textbf{h}'_{p(v,u)}={\rm Encoder}(p(v,u))$\\ $\gamma^\mathcal{P}_{v,u}=\sigma(\textbf{a}^T_{\mathcal{P}}\cdot [\textbf{h}'_v || \textbf{h}'_{p(v,u)}])$\end{tabular} \\ \cline{4-5} 
                                                                                &                     &                                                 & \multicolumn{2}{c|}{$\alpha^\mathcal{P}_{v,u}=\frac{\exp(\gamma^\mathcal{P}_{v,u})}{\sum_{k\in\mathcal{N}^\mathcal{P}_v}\exp(\gamma^\mathcal{P}_{v,k})},\,\,\textbf{z}^\mathcal{P}_v=\sigma(\sum_{k\in\mathcal{N}^\mathcal{P}_v}\alpha^\mathcal{P}_{v,k}\textbf{h}'_{p(v,k)})$}                                                     \\ \hline
\begin{tabular}[c]{@{}c@{}}Semantic\\ fusion\end{tabular}                  & {$\textbf{h}_v=\sum_r \textbf{z}_v^r+W_0\textbf{x}_v$}                  & {\begin{tabular}[c]{@{}c@{}}$\alpha_v^t=\frac{\exp{({\rm LeakyReLU}(\textbf{a}^T[\textbf{z}_v^t || \textbf{h}'_v]))}}{\sum_{k} \exp{({\rm LeakyReLU}(\textbf{a}^T[\textbf{z}_v^k || \textbf{h}'_v]))}}$\\ $\textbf{h}_v=\alpha_v\textbf{h}'_v+\sum_{t}\alpha_v^t\textbf{z}^t_v$\end{tabular}} & \multicolumn{2}{c|}{\begin{tabular}[c]{@{}c@{}}$w_{\mathcal{P}}=\frac{1}{||V^{\phi(v)}||}\sum_{k\in V^{\phi(v)}}\textbf{q}^T\cdot\tanh(\textbf{W}\textbf{z}_k^\mathcal{P}+\textbf{b})$\\ $\beta_{\mathcal{P}_i}=\frac{\exp(w_{P_i})}{\sum_{\mathcal{P}_j}\exp(w_{P_j})},\,\,\textbf{h}_v = \sum_{\mathcal{P}_i}\beta_{\mathcal{P}_i}\textbf{z}^{\mathcal{P}_i}_v$\end{tabular}}       \\ \hline
\end{tabular}
\caption{The unified framework of existing metapath-based HGNNs.}
\label{tab:stages}
\end{table*}

In our re-implemention experiments of HGB, we find that the attention values are mainly dominated by the edge-type embeddings. This is revealed from the observation that attention values within each relation are similar, while those among different relations ary significantly. Figure~\ref{tab:stdev} (a) depicts this observation and Figure~\ref{tab:stdev} (b) shows the statistics of standard deviations of attention values within each relation and among all relations for each target node on the ACM dataset.

This observation motivates us to investigate the necessity of neighbor attention within each relation and semantic attention among different relations in HGNNs.

\subsection{Framework of existing metapath-based methods}
For each layer of existing metapath-based methods, the calculation can be divided into three primary steps, \textit{feature projection}, \textit{neighbor aggregation}, and \textit{semantic fusion}, as presented in Table~\ref{tab:stages}. The feature projection step aims to map raw feature vectors of different types of nodes into the same data space, usually presented as one linear projection layer. Then the neighbor aggregation step aggregates feature vectors of neighbors for each semantic scope, e.g., for each relation (can be viewed as 1-hop metapath) in RGCN, for each node type in HetGNN, or for each metapath in HAN and MAGNN. Finally, the semantic aggregation step fuses those semantic vectors across all semantics and outputs the final embedding for each node.

After removing neighbor attention, since both the one-layer feature projection and neighbor aggregation contain no nonlinear functions, the order of the two steps can be exchanged. Further, as the neighbor aggregation step does not involve any trainable parameters, it can be put ahead in the pre-processing step, and its results are shared across all training epochs.

In later experiments, we find a multi-layer block for feature projection can further enhance the performance. Therefore, SeHGNN employs an MLP block rather than a single linear layer for each metapath in the feature projection step.

\subsection{Experiment settings}

The evaluation of SeHGNN involves four medium-scale datasets from HGB benchmark~\cite{HGB} and a large-scale dataset ogbn-mag\footnote{https://ogb.stanford.edu/docs/nodeprop/\#ogbn-mag} from the OGB challenge~\cite{hu2021ogb}. The statistics of these heterogeneous graphs are summarized in Table~\ref{tab:dataset}.
For the medium-scale datasets, we follow the dataset configuration requirements of the HGB benchmark. Specifically, the target type nodes are split with 30\% for local training and 70\% for online test, where labels of test nodes are not made public and researchers have to submit their predictions to the website of HGB benchmark for online evaluation. For local training, we randomly split the 30\% nodes into 24\% for training for 6\% for validation. We compare our results with the baseline scores reported in the HGB paper, and all scores are the average of 5 different local data partitions. For the ogbn-mag dataset, we follows the official data partition where papers published before 2018, in 2018, and since 2019 are nodes for training, validation, and test, respectively. We compare our results with the scores on the OGB leaderboard, and all scores are the average of 10 separate trainings.

\begin{table}[!t]
\tabcolsep=4pt
\centering
\begin{tabular}{ccccc}
\hline
Dataset  & \#Nodes   & \begin{tabular}[c]{@{}c@{}}\#Node\\ types\end{tabular} & \#Edges    & \#Classes \\ \hline
DBLP     & 26,128    & 4 & 239,566    & 4   \\
ACM      & 10,942    & 4 & 547,872    & 3   \\
IMDB     & 21,420    & 4 & 86,642     & 5   \\
Freebase & 180,098   & 8 & 1,057,688  & 7   \\
Ogbn-mag & 1,939,743 & 4 & 21,111,007 & 349 \\ \hline
\end{tabular}
\caption{The statistics of datasets used in this paper.} \label{tab:dataset}
\end{table}

\begin{table}[!t]
\tabcolsep=2pt
\centering
\begin{tabular}{ccccc}
\hline
                     & \multicolumn{2}{c}{Feature propagation} & \multicolumn{2}{c}{Label Propagation} \\ \hline
\multicolumn{1}{l}{} & Max Hop & \#Metapaths         & Max Hop & \#Metapaths       \\ \hline
DBLP                 & 2                  & 5                   & 4                 & 4                 \\
IMDB                 & 4                  & 25                  & 4                 & 12                \\
ACM                  & 4                  & 41                  & 3                 & 9                 \\
Freebase             & 2                  & 73                  & 3                 & 9                 \\
Ogbn-mag             & 2                  & 10                  & 2                 & 5                 \\ \hline
\end{tabular}
\caption{The maximum hops and numbers of metapaths for raw feature propagation and label propagation.} \label{tab:metapaths}
\end{table}

We adopt a simple metapath selection method where we preset the maximum hop and use all available metapaths no more than this maximum hop. We test different combinations of maximum hops for raw feature propagation and label propagation and the final choices are listed in Figure~\ref{tab:metapaths}.

For all experiments, SeHGNN adopts a two-layer MLP for each metapath in the feature projection step, where the dimension of hidden vectors is 512. In the transformer-based fusion module, the dimension of query and key vectors are 1/4 of hidden vectors, the dimension of value vectors equals that of hidden vectors, and the number of heads is 1.

SeHGNN is optimized with Adam \cite{adam} during training. The learning rate is 0.0003 and the weight decay is 0.0001 for the Freebase dataset, and the learning rate is 0.001 and the weight decay is 0 for others.

\subsection{Further analysis of time complexity}

\begin{table}[!t]
\centering
\tabcolsep=3.5pt
\begin{tabular}{ccccc}
\hline
       & \begin{tabular}[c]{@{}c@{}}Feature\\ projection\end{tabular} & \begin{tabular}[c]{@{}c@{}}Neighbor\\ aggregation\end{tabular} & \begin{tabular}[c]{@{}c@{}}Semantic\\ fusion\end{tabular} \\ \hline
HAN    & $O((n{+}m)d^2)$ & $O(nke_1d{+}(n{+}m)kd)$ & $O(nkd^2)$  \\
HGB    & $O((n{+}m)ld^2)$ & \multicolumn{2}{c}{\,\,\,\,\,\,\,\,\,\,\,\,\,\,\,\,$O(ne_2d{+}(n{+}m)ld)$} \\ \hline
\end{tabular}
\caption{Revised time complexity on HAN and HGB with consideration of nodes of all types.}
\label{tab:time_complexity2}
\end{table}

The computation complexity of HAN and HGB methods is challenging to estimate as it encompasses various node types and the reuse of projected neighbor node features in the neighbor aggregation step. To make a straightforward comparison, Table 5 includes the complexity of linear projection for target type nodes in HAN and HGB but omits that of nodes of other types. As illustrated in Table 9, we add the computation complexity of nodes of other types assuming the number of these nodes involved in one training batch to be $m$. These additions include the linear projection in the feature projection step and the calculation of neighbor attentions in the neighbor aggregation step.

When training in full-batch mode, $n$ and $m$ represent the total number of target type nodes and other type nodes in the heterogeneous graph, respectively. If $n$ and $m$ are comparable, i.e., $O(n)=O(m)$, then results in Table 9 can be reduced to those in Table 5.

However, for mini-batch training, $m$ can be significantly larger than $n$ for each training batch. In the case of a small batch size $n$ and almost no reusable projected neighbor node features, $m$ increases exponentially with the metapath length or number of layers (i.e., $O(m)=O(nke_1)$ for HAN and $O(m)=O(ne_2)$ for HGB). In this scenario, the computation complexity results on HAN and HGB in Table 5 show a significant under-estimation.

In contrast, estimating the computation complexity of SeHGNN is much simpler. Because the neighbor aggregation is calculated in the pre-processing step, eliminating the involvement of nodes of other types in each training batch.

\end{document}